\documentclass[10pt,twocolumn,letterpaper]{article}

\usepackage[pagenumbers]{cvpr}

\usepackage{latexsym}
\usepackage[hyphens]{url}  
\usepackage[utf8]{inputenc}
\usepackage{helvet}  
\usepackage{courier}  
\usepackage{graphicx} 
\usepackage{natbib}  
\usepackage{caption} 
\usepackage{algorithm}
\usepackage{algpseudocode}
\usepackage[T1]{fontenc}    
\usepackage{booktabs}       
\usepackage{amsfonts}       
\usepackage{nicefrac}       
\usepackage{enumitem}
\usepackage{xcolor}         
\usepackage{bm}
\usepackage{amsmath,mathrsfs,amssymb}
\usepackage{longtable}
\usepackage{threeparttable}
\usepackage{multirow}
\usepackage{multicol}
\usepackage{soul}
\usepackage[american]{babel}
\usepackage{lipsum}
\usepackage{subcaption}
\usepackage{makecell}   
\usepackage{tabularx}
\allowdisplaybreaks[3]

\definecolor{cvprblue}{rgb}{0.21,0.49,0.74}
\usepackage[pagebackref,breaklinks,colorlinks,allcolors=cvprblue]{hyperref}

\def\confName{CVPR}
\def\confYear{2026}

\title{R2Q: Towards Robust 2-Bit Large Language Models via Residual Refinement Quantization}

\author{
    \textbf{Jiayi Chen}\textsuperscript{1} \quad
    \textbf{Jieqi Shi}\textsuperscript{1,}\thanks{Corresponding author.} \quad
    \textbf{Jing Huo}\textsuperscript{1} \quad
    \textbf{Chen Wu}\textsuperscript{2} \\[5pt]
    \textsuperscript{1}Nanjing University \quad
    \textsuperscript{2}Microsoft AI
}
\begin{document}
\maketitle
\begin{abstract}
The rapid progress of Large Language Models (LLMs) has brought substantial computational and memory demands, spurring the adoption of low-bit quantization. While 8-bit and 4-bit formats have become prevalent, extending quantization to 2 bits remains challenging due to severe accuracy degradation. To address this, we propose Residual Refinement Quantization (R2Q)—a novel 2-bit quantization framework that decomposes the process into two sequential 1-bit sub-quantizations, forming an adaptive quantization lattice. Extensive evaluations on Llama, OPT, and Qwen across diverse benchmarks—covering question answering, commonsense reasoning, and language modeling—demonstrate that R2Q consistently outperforms existing 2-bit quantization methods in both fine-grained and coarse-grained settings. By refining quantization through a residual learning mechanism, R2Q enhances performance, improves training stability, and accelerates convergence under extreme compression. Furthermore, its modular design enables seamless integration with existing quantization-aware training (QAT) frameworks.
\end{abstract}    
\section{Introduction}\label{sec:introduction}
Large Language Models (LLMs) have radically advanced natural language processing, achieving strong performance on tasks ranging from text generation to complex reasoning. Yet their massive parameter scales and substantial computational costs introduce significant deployment challenges. To mitigate these burdens, recent research has increasingly turned to more efficient numerical formats—initially FP16~\citep{micikevicius2017fp16} and BFloat16 (BF16)~\citep{kalamkar2019bf16}, and more recently low-bit quantization, particularly 8-bit and 4-bit. These approaches, including integer~\citep{dettmers2022llmint8,yao2022zeroquant,xi2023int4}, floating-point~\citep{micikevicius2022fp8,liu2023fp4}, and NormalFloat~\citep{dettmers2023qlora} formats, have proven effective in reducing memory footprint and compute cost while preserving strong model quality. Despite these advances, the stringent memory budgets of resource-constrained edge devices—such as smartphones, UAVs~\citep{mohsan2023UAV}, and AR/VR platforms~\citep{arena2022AR,MetaVR}—demand even more aggressive compression.

\begin{figure}[!htb]
    \centering
    \includegraphics[width=1.0\linewidth,trim=3.8cm 3cm 3cm 2cm,clip]{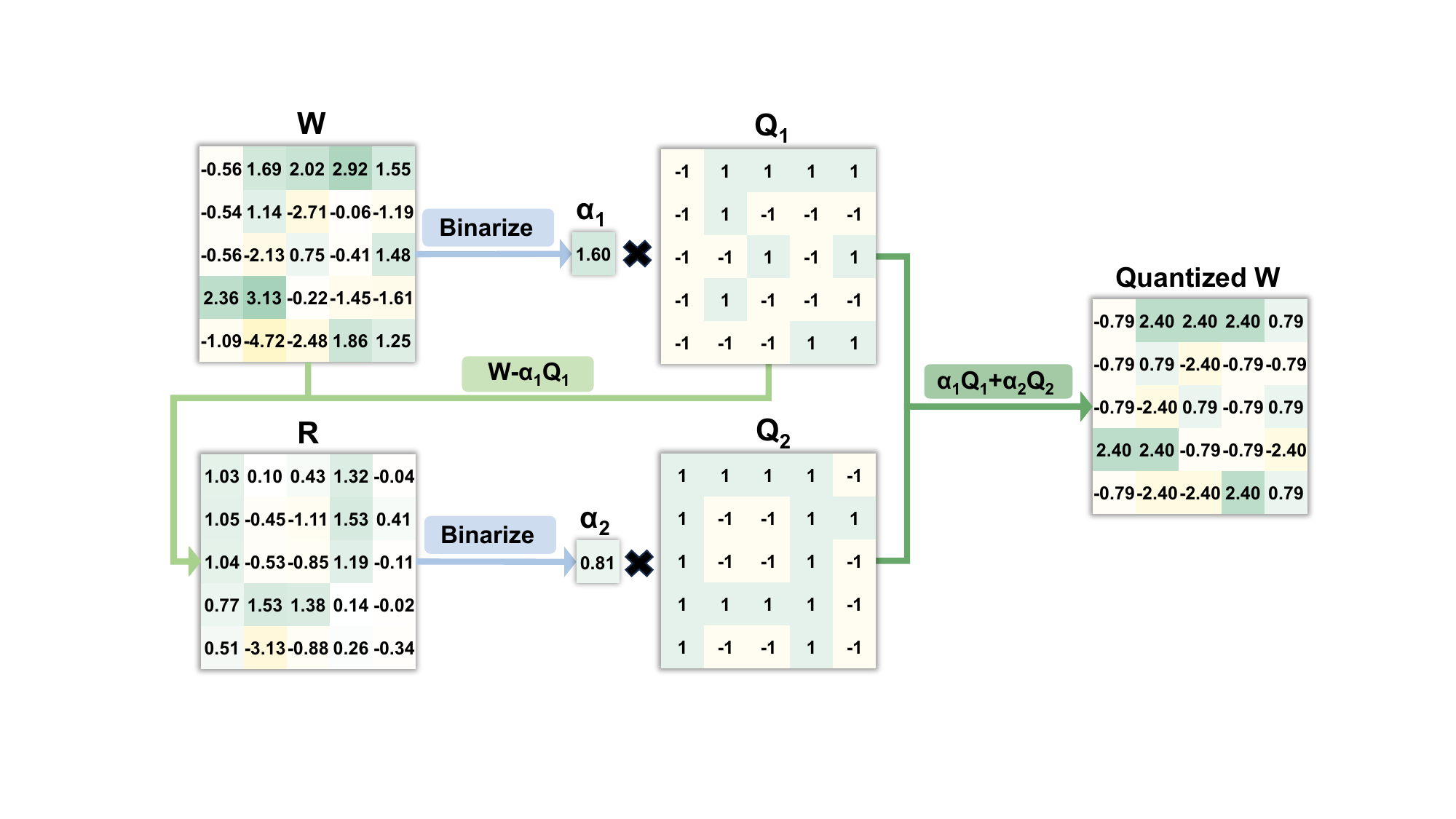}
    \caption{Overview of the Residual Refinement Quantization (R2Q) mechanism. The full-precision weight $\textbf{W}$ and the first-step residual $\textbf{R}$ are binarized into two 1-bit kernels, $\textbf{Q}_1$ and $\textbf{Q}_2$, which are merged to reconstruct $\textbf{W}$.}
    \label{fig:flow}
\end{figure}

Binary quantization represents the most extreme form of compression, eliminating costly multiplications altogether~\citep{rastegari2016xnor}, a benefit unattainable for 2-bit and higher-bit formats. However, achieving competitive accuracy typically requires training from scratch~\citep{wang:bitnet,wang2025ifairy} or heavy fine-tuning~\citep{xu2024onebit}. Ternary quantization~\citep{chen2024ternaryllm,ma:1.58bitLLM,yuan:vit1.58b} adds a zero level to improve expressiveness but remains limited by hardware primarily optimized for binary operations. Positioned between these extremes, 2-bit quantization offers a trade-off, combining high compression with broad hardware compatibility. The key challenge lies in constructing effective mapping strategies. Existing methods~\citep{chee2023quip,du2024bitdistiller,shao2023omniquant} predominantly follow integer quantization pipelines, where high-precision weights are scaled, shifted, and rounded (RTN). Yet due to the restricted representational capacity of 2-bit formats, maintaining accuracy often requires fine-grained grouping~\citep{shao2023omniquant,du2024bitdistiller}, limiting their practicality.

To overcome these challenges, we introduce \textbf{Residual Refinement Quantization (R2Q)}—a novel quantization paradigm specifically designed for 2-bit settings. As illustrated in Figure~\ref{fig:flow}, R2Q decomposes the 2-bit quantization process into two sequential 1-bit subproblems: the first provides a coarse approximation of the original weights, while the second refines the residual. This decomposition yields a \textbf{learnable and flexible} quantization lattice, allowing R2Q to adapt to the intrinsic distribution of model weights. In contrast to RTN’s static and uniform lattice, R2Q’s adaptive structure produces a more uniform and accurate parameter mapping (see Figure~\ref{fig:int-residual}), leading to more \textbf{stable and faster-converging} Quantization-Aware Training (QAT). Finally, R2Q’s \textbf{modular} design enables straightforward integration into existing QAT frameworks.

Our main contributions are summarized as follows:
\begin{itemize}
    \item We propose R2Q, a novel 2-bit quantization strategy that decomposes the 2-bit quantization problem for LLMs into two simpler 1-bit sub-problems, enabling a more flexible and learnable quantization lattice.
    
    \item We designed R2Q as a modular and integrable solution that can be readily integrated into existing QAT frameworks, offering substantial and consistent improvements in stability and convergence efficiency.
    
    \item We validate R2Q through extensive experiments on multiple benchmarks, demonstrating that it outperforms existing methods under 2-bit quantization, especially when combined with coarse-grained group quantization.
\end{itemize}

The remainder of the paper is organized as follows: Section~\ref{sec:related work} reviews related work. Section~\ref{sec:background} introduces preliminaries. Section~\ref{sec:Methodology} presents the R2Q framework. Section~\ref{sec:Experiments} conducts experiments. Section~\ref{sec:Discussion_and_Future_Work} discusses results and future directions.

\section{Related Work}\label{sec:related work} 
This section reviews related work on LLM quantization. We organize prior methods along two dimensions: \textit{quantization formats} and \textit{quantization strategies}.

\subsection{Quantization Format} 
A straightforward approach to model compression is reducing fractional and exponent precision, as in FP16 and BF16. BF16, with improved numerical stability, is now standard in state-of-the-art models such as OpenPangu, Llama~\citep{dubey2024llama3}, and Qwen~\citep{bai2023qwen}. However, growing model scales demand more aggressive quantization. At its core, LLM quantization maps continuous values to a finite discrete set for efficient compression.

For 8-bit formats, 8-bit integer (INT8) and 8-bit floating-point (FP8) are prominent. LLM.int8()~\citep{dettmers2022llmint8} successfully implemented INT8 quantization for LLMs by using vector-wise quantization and a special mixed-precision decomposition. Concurrently, FP8~\citep{micikevicius2022fp8}, such as E4M3 and E5M2, were proven effective for both training and inference by matching the performance of 16-bit formats without requiring hyperparameter changes.

In the 4-bit regime, INT4 methods such as~\citet{xi2023int4} mitigate activation outliers via Hadamard transforms and importance sampling. FP4~\citep{liu2023fp4} improves robustness with optimized quantization parameters and exponent shifting. QLoRA~\citep{dettmers2023qlora} further enables efficient fine-tuning by introducing the NormalFloat4 (NF4) format with LoRA and double quantization.

Binary and ternary quantization typically rely on the indicator function~\citep{wang:bitnet,xu2024onebit} and the sign function~\citep{chen2024ternaryllm,ma:1.58bitLLM,yuan:vit1.58b}, respectively. The optimality of these mappings has been established under specific distributional assumptions~\citep{rastegari2016xnor,li2016tnn}. In contrast, most existing 2-bit approaches adopt integer-based formats with RTN~\citep{chee2023quip,du2024bitdistiller,shao2023omniquant}, a design inherited from higher-bit quantization but poorly aligned with the limited representational capacity of only four discrete values. Recent work~\citep{wang2025ifairy} explores weights in the complex domain, yet this direction has not been scaled to larger models.

\subsection{Quantization Strategies}
Quantization strategies can be grouped into two primary categories: Post-Training Quantization (PTQ) and Quantization-Aware Training (QAT).

PTQ applies quantization without retraining, often using calibration data. GPTQ~\citep{frantar2022gptq} leverages second-order approximations of weight sensitivity via the Hessian matrix to achieve accurate 3/4-bit quantization. AWQ~\citep{lin2024awq} and SmoothQuant~\citep{xiao2023smoothquant} focus on activation-aware weight scaling to preserve salient channel representations. QuaRot~\citep{ashkboos2024quarot} and FlatQuant~\citep{sun2024flatquant} simplify quantization by flattening weight or activation distributions. While efficient, PTQ methods degrade sharply under extreme bit-widths (e.g., 2-bit).

QAT integrates quantization into training or fine-tuning. Although more resource-intensive, QAT delivers superior results in ultra-low-bit settings. QLoRA~\citep{dettmers2023qlora} combines 4-bit quantization with LoRA, leveraging NF4 and double quantization to minimize memory usage. LLM-QAT~\citep{liu2023llm-qat} employs data-free distillation from a full-precision teacher, while BitDistiller~\citep{du2024bitdistiller} introduces a confidence-aware KL divergence objective tailored for 2-bit quantization. \citet{lee2025unifying} propose a progressive approach that first applies PTQ to obtain a 4-bit model, followed by QAT with knowledge distillation to further reduce precision to 2 bits. Following this line of work, we adopt a distillation-based QAT framework.

\begin{figure}[!htb]
    \centering
    \includegraphics[width=0.9\linewidth,trim=7.5cm 3.5cm 8.5cm 3cm,clip]{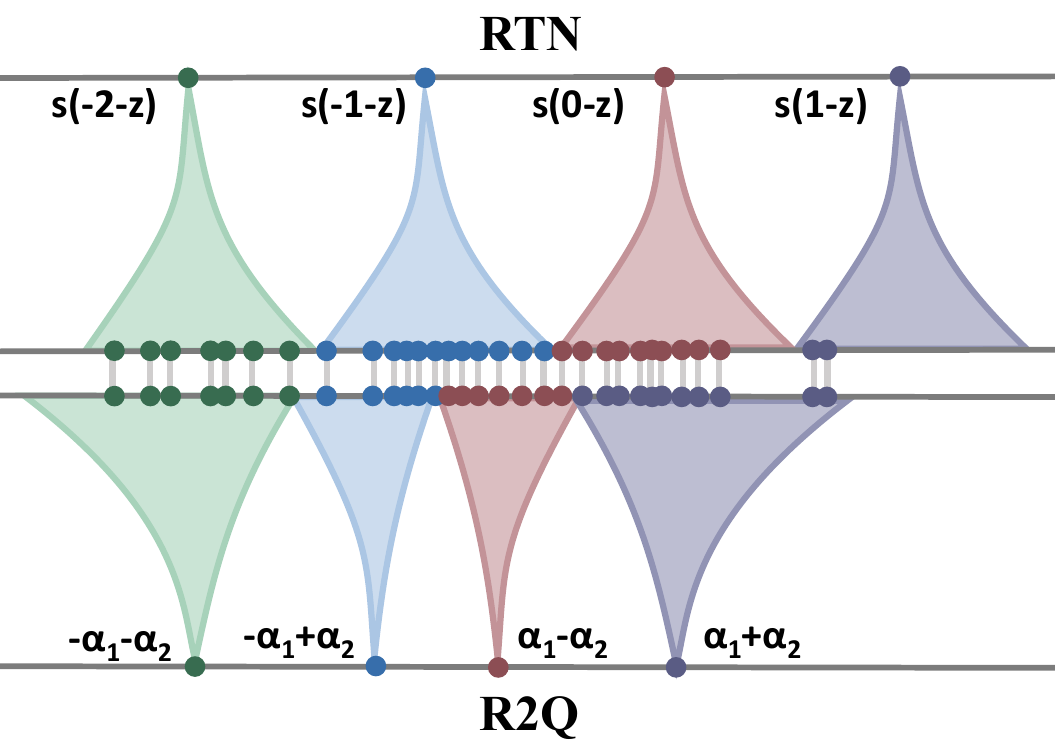}
    \caption{Comparison of RTN (top) and R2Q (bottom) in 2-bit quantization. Gray lines connect points that correspond to the same real value. Points within the base of the conical region are mapped to the quantized position indicated by the apex. We refer to this structure as a \textbf{quantization lattice}. $s$ and $z$ are the scaling parameters and zero points for RTN, while $\alpha_1$ and $\alpha_2$ represent the scaling parameters of the two kernels of R2Q. respectively. R2Q achieves an \textbf{adaptive} mapping for imbalanced data distributions, whereas RTN results in fixed and uniform allocation—as illustrated in the top image, where only two real values fall into the purple lattice—leading to inefficient use of the limited four quantization levels available in 2-bit quantization.}
    \label{fig:int-residual}
\end{figure}
\section{Preliminaries}\label{sec:background}

This section introduces the preliminaries of LLM quantization, covering quantization and dequantization, group-wise quantization, and static lattice mapping.

\subsection{Problem Formulation}
LLMs are predominantly decoder-based Transformers, whose substantial memory and computational costs arise mainly from their linear layers. Consequently, quantization plays a crucial role in enabling efficient deployment.

We focus our analysis on a single linear layer, as the same process applies to all others. The operation is defined as
\begin{equation}
f(\mathbf{X}) = \mathbf{W}\boldsymbol{x},
\end{equation}
where $\boldsymbol{x} \in \mathbb{R}^{D_{in}}$ denotes the input vector and $\mathbf{W} \in \mathbb{R}^{D_{out} \times D_{in}}$ represents the full-precision weight matrix.

Quantization maps continuous values in $\mathbf{W}$ to a discrete set of points from a predefined codebook $\mathcal{C} = \{c_1, c_2, \ldots, c_{2^k}\}$, where $k$ denotes the bit width. The process can be expressed as
\begin{equation}
\mathbf{Q} = \mathcal{Q}_\theta(\mathbf{W}),
\end{equation}
where $\mathbf{Q}$ is the quantized weight matrix with each entry $Q_{ij} \in \mathcal{C}$, and $\mathcal{Q}_\theta$ denotes the quantization function parameterized by $\theta$ (e.g., scaling factors and zero-points).
The approximate reconstruction of $\mathbf{W}$ is obtained through a dequantization function $\mathcal{D}_\theta$:
\begin{equation}
\hat{\mathbf{W}} = \mathcal{D}_\theta(\mathbf{Q}) \approx \mathbf{W}.
\end{equation}

\subsection{Group-Wise Quantization}
Group-wise quantization is widely adopted in modern quantization methods~\citep{frantar2022gptq,shao2023omniquant,liu2023llm-qat,du2024bitdistiller} to enhance precision and reduce quantization error.
This approach partitions the weight matrix $\mathbf{W}$ into smaller disjoint groups, allowing quantization to operate at a finer granularity. Specifically, $\mathbf{W}$ is divided into $N$ groups, which can be expressed as
\begin{equation}
\mathbf{W} = [\boldsymbol{w}^{(1)}, \boldsymbol{w}^{(2)}, \ldots, \boldsymbol{w}^{(N)}], \quad \boldsymbol{w}^{(i)} \in \mathbb{R}^{G},
\end{equation}
where $G$ is the group size, and each $\boldsymbol{w}^{(i)}$ denotes a contiguous subset of weights.
Each group is independently quantized and dequantized via group-specific parameters
\begin{equation}
\hat{\boldsymbol{w}}^{(i)} =\mathcal{D}_{\theta^{(i)}}\left( \mathcal{Q}_{\theta^{(i)}}\left(\boldsymbol{w}^{(i)}\right)\right), \quad \hat{\boldsymbol{w}}^{(i)} \in \mathcal{C}^{G},
\end{equation}
The quantized matrix $\hat{\mathbf{W}}$ is then reconstructed by concatenating the quantized groups
\begin{equation}
\hat{\mathbf{W}} = [\hat{\boldsymbol{w}}^{(1)}, \hat{\boldsymbol{w}}^{(2)}, \ldots, \hat{\boldsymbol{w}}^{(N)}].
\end{equation}
This group-wise strategy enables finer adaptation to local weight distributions and mitigates quantization errors, particularly in low-bit quantization scenarios.

\subsection{Static Lattice Mapping}\label{sec:static lattice mapping}
Round-to-Nearest (RTN) quantization, which relies on a static lattice, is one of the most commonly used strategies in 2-bit quantization~\citep{du2024bitdistiller,chee2023quip}. RTN scales and shifts a fixed set of integer base points while preserving uniform spacing between quantization levels.

In RTN, an affine transformation maps real-valued weights to a discrete signed (or unsigned) integer range.
For the signed case, the target range is $[q_{\min}, q_{\max}] = [-(2^{k-1}), 2^{k-1}-1]$.
Given a weight group with real-valued range $r_{\min}^{(i)} = \min(\boldsymbol{w}^{(i)})$ and $r_{\max}^{(i)} = \max(\boldsymbol{w}^{(i)})$, quantization is defined by a scaling factor $s$ and a zero-point $z$, which can be fomulated as

\begin{equation}
\begin{aligned}
s^{(i)} &= \frac{r_{\max}^{(i)} - r_{\min}^{(i)}}{q_{\max} - q_{\min}}\ , \\
\boldsymbol{z}^{(i)} &= \left\lfloor\frac{q_{\text{min}} - r_{\min}^{(i)}}{s^{(i)}}\right\rceil\ , \\
\boldsymbol{q}^{(i)} = \mathcal{Q}_{s^{(i)},z^{(i)}}\left(\boldsymbol{w}^{(i)}\right)& 
= \text{clip}\left( \left\lfloor\frac{\boldsymbol{w}^{(i)}}{s^{(i)}}\right\rceil + z^{(i)}, \, q_{\min}, \, q_{\max} \right)\ , \label{eq:quantize}
\end{aligned}
\end{equation}
where $\lfloor\cdot\rceil$ denotes rounding to the nearest integer, and $\text{clip}(\cdot)$ ensures that quantized values remain within the target range.
The dequantized weight is then recovered as
\begin{equation}
    \hat{\boldsymbol{w}}^{(i)} = \mathcal{D}_{s^{(i)},z^{(i)}}\left(\boldsymbol{q}^{(i)}\right)=
    s^{(i)} \cdot (\boldsymbol{q}^{(i)} - z^{(i)})\ .
\end{equation}
While RTN is effective for 4-bit or higher precision, where the number of quantization levels sufficiently captures non-uniform distributions~\citep{dettmers2022llmint8,xi2023int4,frantar2022gptq}, its uniform lattice is poorly suited for 2-bit quantization.
With only four quantization levels, LLM weight distributions—typically non-uniform and heavily centered—cause most values to cluster near central points, leaving few outliers to occupy other levels (Figure~\ref{fig:int-residual}).
This imbalance wastes limited representation capacity and reduces parameter distinctiveness, underscoring the need for a more adaptive and flexible lattice design.

\section{Methodology}\label{sec:Methodology}
Conventional static lattice quantization, as discussed in Section~\ref{sec:background}, faces substantial challenges under extreme bit-width constraints (e.g., 2-bit). To address this, we propose Residual Refinement Quantization (R2Q), a novel 2-bit quantization framework that decomposes the quantization task into two successive 1-bit subproblems. This section first reformulates the 2-bit quantization problem, derives the optimal solution for each 1-bit subproblem, and then presents the complete R2Q algorithm.

\subsection{2-bit Quantization via 1-bit Decomposition}\label{sec:2bit decompose}
We start by decomposing the 2-bit quantization problem. A 2-bit quantization codebook contains four discrete values, i.e., $\mathcal{C} = \{c_1, c_2, c_3, c_4\}$. Directly optimizing a static lattice by global scaling and shifting, as described in Section~\ref{sec:background}, is non-trivial and often results in large approximation errors.

To overcome this limitation, we reformulate the 2-bit problem as two consecutive 1-bit subproblems. For the $i$-th weight group, the quantized vector $\boldsymbol{w}^{(i)}$ can be approximated as
\begin{equation}\label{eq:2-bit-decomposition}
\boldsymbol{w}^{(i)} \approx \hat{\boldsymbol{w}}^{(i)} = \alpha^{(i)}_1 \boldsymbol{q}^{(i)}_1 + \alpha^{(i)}_2 \boldsymbol{q}^{(i)}_2,
\end{equation}
where $\boldsymbol{q}_1^{(i)}$ and $\boldsymbol{q}_2^{(i)}$ are vectors with elements in $\{+1, -1\}^G$. Here, the first kernel $\alpha_1^{(i)}\boldsymbol{q}_1^{(i)}$ serves as a coarse approximation of $\boldsymbol{w}^{(i)}$, while the second kernel $\alpha_2^{(i)}\boldsymbol{q}_2^{(i)}$ captures the residual for further refinement. The coefficients $\alpha_1^{(i)}$ and $\alpha_2^{(i)}$ are the corresponding scaling factors. 

This decomposition yields an adaptive codebook defined as $\{-\alpha_1^{(i)} - \alpha_2^{(i)},\ -\alpha_1^{(i)} + \alpha_2^{(i)},\ \alpha_1^{(i)} - \alpha_2^{(i)},\ \alpha_1^{(i)} + \alpha_2^{(i)}\}$. The scaling factors $\alpha_1^{(i)}$ and $\alpha_2^{(i)}$ dynamically adjust the lattice scale, enabling greater flexibility in representing diverse weight distributions.

\subsection{Optimal Solution for the 1-bit Subproblem}
For a given vector $\boldsymbol{w}$, we seek an optimal binary approximation parameterized by a direction vector $\boldsymbol{q} \in \{-1, 1\}^{G}$ and a positive scaling factor $\alpha \in \mathbb{R}^+$. This can be formulated as the following optimization problem~\citep{rastegari2016xnor},
\begin{equation}\label{eq:1-bit_subproblem}
\alpha^*, \boldsymbol{q}^* = \underset{\alpha > 0, \mathbf{Q}}{\operatorname{argmin}} \|\boldsymbol{w} - \alpha\boldsymbol{q}\|_F^2 \ , \quad \text{s.t.} \quad q_{i} \in \{-1, 1\} \ .
\end{equation}
Expanding the objective yields
\begin{equation}\label{eq:constraint problem}
\|\boldsymbol{w} - \alpha\boldsymbol{q}\|_F^2 = \|\boldsymbol{w}\|_F^2 - 2\alpha \langle \boldsymbol{w}, \boldsymbol{q} \rangle_F + \alpha^2 \|\boldsymbol{q}\|_F^2 \ ,
\end{equation}
where $\langle \cdot, \cdot \rangle_F$ is the Frobenius inner product. Since $q_{i} \in \{-1, 1\}$, it follows $q_{i}^2=1$, and thus $\|\boldsymbol{q}\|_F^2 = \sum_{i} q_{i}^2 = G$. The objective becomes $\|\boldsymbol{w}\|_F^2 - 2\alpha\sum_{i} w_{i}q_{i} + G\alpha^2$. We can solve for $\alpha$ and $\boldsymbol{q}$ iteratively.For a fixed $\alpha > 0$, minimizing the objective requires maximizing $\sum_i w_i q_i$, which is achieved when $q_i$ shares the same sign as $w_i$. Hence, the optimal binary direction is
\begin{equation}\label{eq:1-bit_solution_Q}
\boldsymbol{q}^* = \mathcal{H}(\boldsymbol{w})\ ,
\end{equation}
\begin{equation}
    \mathcal{H}(w)=\left\{
    \begin{aligned}
        +1\ \ \ \ w\geq0 \ ,\\
        -1\ \ \ \ w<0 \ ,
    \end{aligned}
    \right.
\end{equation}
where the $\mathcal{H}(\cdot)$ function is applied element-wise. Substituting $\boldsymbol{q}^*$ back into the objective and minimizing with respect to $\alpha$, we obtain
\begin{equation}
\begin{aligned}
&\frac{\partial}{\partial \alpha} \left( \|\boldsymbol{w}\|_F^2 - 2\alpha \langle \boldsymbol{w}, \boldsymbol{q}^* \rangle_F + G\alpha^2 \right) = -2\langle \boldsymbol{w}, \boldsymbol{q}^* \rangle_F + 2G\alpha = 0 \ .
\end{aligned}
\end{equation}
Solving for $\alpha$ yields the optimal scaling factor
\begin{equation}\label{eq:1-bit_solution_alpha}
\begin{aligned}
\alpha^* &= \frac{\langle \boldsymbol{w}, \boldsymbol{q}^* \rangle_F}{G} = \frac{\sum_{i} w_{i}\operatorname{sign}(w_{i})}{G} = \frac{\sum_{i} |w_{i}|}{G} = \frac{\|\boldsymbol{w}\|_{\ell1}}{G} \ ,
\end{aligned}
\end{equation}
where $\|\cdot\|_{\ell1}$ denotes the $\ell_1$-norm. So far, we have derived the optimal solution to the 1-bit subproblem. Notably, it does not rely on specific weight distribution—a property not shared by ternary or higher-bit quantization~\citep{li2016tnn}. This characteristic contributes to the improved robustness of R2Q.

\subsection{Residual Refinement Quantization}\label{sec:R2Q}
R2Q employs a two-step quantization strategy: the first step captures a coarse binary approximation, while the second refines the residual. Formally, for each weight group $\boldsymbol{w}^{(i)}$:

\begin{enumerate}
\item Compute $(\boldsymbol{q}_1^{(i)}, \alpha_1^{(i)})$ by minimizing $\|\boldsymbol{w}^{(i)} - \alpha_1^{(i)} \boldsymbol{q}_1^{(i)}\|_F^2$;
\item Compute the residual $\boldsymbol{r}^{(i)} = \boldsymbol{w}^{(i)} - \alpha_1^{(i)}\boldsymbol{q}_1^{(i)}$, and obtain $(\boldsymbol{q}_2^{(i)}, \alpha_2^{(i)})$ by minimizing $|\boldsymbol{r}^{(i)} - \alpha_2^{(i)} \boldsymbol{q}_2^{(i)}|_F^2$.
\end{enumerate}

First, a coarse 1-bit approximation is found by solving the optimization subproblem for the original weights $\boldsymbol{w}^{(i)}$, Using the optimal solutions derived in Eq.~\eqref{eq:1-bit_solution_Q} and Eq.~\eqref{eq:1-bit_solution_alpha}
\begin{align}
\boldsymbol{q}_1^{(i)} &= \mathcal{H}(\boldsymbol{w}^{(i)})\ ,  \\
\alpha_1^{(i)} &= \frac{\|\boldsymbol{w}^{(i)}\|_{\ell1}}{G}\ .
\end{align}
This results in a residual error $\boldsymbol{r}^{(i)}$ which captures the information lost in the coarse approximation
\begin{equation}
\boldsymbol{r}^{(i)} = \boldsymbol{w}^{(i)} - \alpha_1^{(i)}\boldsymbol{q}_1^{(i)} \ . 
\end{equation}
Next, to refine the approximation, a second 1-bit quantization is performed on this residual error $\boldsymbol{r}^{(i)}$. This step solves for the refinement term by applying the same strategy
\begin{align}
\boldsymbol{q}_2^{(i)} &= \mathcal{H}(\boldsymbol{r}^{(i)}) \ ,  \\
\alpha_2^{(i)} &= \frac{\|\boldsymbol{r}^{(i)}\|_{\ell1}}{G} \ . 
\end{align}
Finally, the full 2-bit quantized approximation of the weight group, $\hat{\boldsymbol{w}}^{(i)}$, is obtained by combining the coarse approximation and the refined residual term, as defined in Eq.~\eqref{eq:2-bit-decomposition}:
\begin{equation}
\hat{\boldsymbol{w}}^{(i)} = \alpha_1^{(i)}\boldsymbol{q}_1^{(i)} + \alpha_2^{(i)}\boldsymbol{q}_2^{(i)} \ . 
\end{equation}
This reconstructed $\hat{\boldsymbol{w}}^{(i)}$ is used in the forward pass of the network. Our quantization strategy provides an adaptive quantization lattice, facilitating adaptive mapping for ill-shaped weight distributions, as shown in Figure~\ref{fig:int-residual}. R2Q fully exploits the limited 2-bit quantization space via two strategies: first, decomposing the 2-bit quantization problem into a 1-bit coarse estimate and a 1-bit residual refinement; second, ensuring the \textbf{distribution-independent optimal solution} of each 1-bit subproblem. The former enhances lattice utilization, while the latter ensures distribution-agnostic robustness.

\subsection{Backpropagation}
The quantization process in R2Q involves the non-differentiable function $\mathcal{H}(\cdot)$, which blocks gradient flow during backpropagation. To circumvent this issue, we adopt the Straight-Through Estimator (STE)~\citep{bengio2013STE}, which approximates the gradient of the quantization function as the identity mapping. This allows gradients from the loss function $\mathcal{L}$ to propagate back to the full-precision weights $\mathbf{W}$. After R2Q quantizes and reconstructs $\mathbf{W}$, the backward propagation is formally expressed as
\begin{equation}
\frac{\partial \mathcal{L}}{\partial \mathbf{W}} = \frac{\partial \mathcal{L}}{\partial \hat{\mathbf{W}}} \frac{\partial \hat{\mathbf{W}}}{\partial \mathbf{W}} \ .
\end{equation}
With STE, the quantizer’s gradient is approximated as
\begin{equation}
\frac{\partial \hat{\mathbf{W}}}{\partial \mathbf{W}} \approx \mathbf{I} \ .
\label{eq:backprop}
\end{equation}
Substituting into Eq.~\eqref{eq:backprop}, we obtain
\begin{equation}
\frac{\partial \mathcal{L}}{\partial \mathbf{W}} \approx \frac{\partial \mathcal{L}}{\partial \hat{\mathbf{W}}} \ .
\end{equation}
This approximation yields a simplified and tractable gradient for updating the full-precision weights.

\section{Experiments}\label{sec:Experiments}
This section presents a comprehensive evaluation designed to validate the effectiveness and robustness of the proposed R2Q method for 2-bit Large Language Model quantization. We begin by delineating the experimental configurations, including the experimental Setup, the data-free training paradigm, and the suite of evaluation benchmarks. Subsequently, we analyze the comparative results, focusing on language understanding, generative modeling, and quantization stability to substantiate the capabilities of R2Q.

\begin{table*}[t]
\centering
\caption{Comparison of R2Q and other 2-bit quantization methods under coarse-grained (group size = -1) and fine-grained (group size = 128) settings. Evaluations are based on language understanding and modeling tasks. CR denotes the compression ratio of the model's memory usage. For ARC-c/e, BoolQ, Hellaswag, PIQA, and Winogrande, we report accuracy. For WikiText-2, we report PPL. R2Q uses two 1-bit kernels, each with its own scaling parameter, resulting in twice the number of scaling parameters under fine-grained quantization. To match the parameter count, we set R2Q's group size to 256 (i.e., 2 × 128).}
{\setlength{\tabcolsep}{2pt}
\resizebox{\linewidth}{!}{ %
\begin{tabular}{ccccccccccccc}
\toprule
\textbf{Model} & \textbf{Bit-Width} & \textbf{Method} & \textbf{Group-Size}&\textbf{CR} & \textbf{ARC-c} & \textbf{ARC-e}$\uparrow$ & \textbf{BoolQ} $\uparrow$ & \textbf{Hella.} $\uparrow$ & \textbf{PIQA}$\uparrow$ & \textbf{Wino.}$\uparrow$& \textbf{MMLU}$\uparrow$ & \textbf{Wiki2}$\downarrow$  \\
\midrule
\multirow{7}{*}{Llama-7B} & bf16 & \textbackslash & \textbackslash & 100.00\% & 41.98 & 75.38 & 75.17 & 56.95 & 78.78 & 69.46 & 31.33 & 9.39 \\
\cmidrule{2-13}
&\multirow{6}{*}{2-bit} & LLM-QAT & -1 & 12.3\% & 19.45 & 26.81 & 37.83 & 25.71 & 51.90 & 50.28 & 22.95 & 1776.98 \\
& & BitDistiller & -1 & 12.3\% & 23.38 & 25.88 & 53.88 & 25.59 & 51.46 & 50.43 & \textbf{24.61} & 15938.61 \\
&  & \textbf{R2Q (ours)} & -1 & 12.3\% & \textbf{27.47} & \textbf{56.82} & \textbf{59.36} & \textbf{44.44} & \textbf{70.08} & \textbf{57.54} & 24.08 & \textbf{17.13} \\
\cmidrule{3-13}
& & LLM-QAT & 128 &15.1\% & 26.19 & 42.00 & 58.32 & 35.32 & 62.13 & 54.22 & 23.91 & 115.52 \\
& & BitDistiller & 128 &15.1\% & \textbf{28.67} & 57.91 & 63.18 & 41.76 & 69.10 & \textbf{60.22} & 24.26 & 33.51 \\
& & \textbf{R2Q (ours)}& 256 &15.0\% & 28.24 & \textbf{59.18} & \textbf{64.89} & \textbf{45.37} & \textbf{70.62} & 59.59 & \textbf{24.28} & \textbf{16.96} \\
\midrule
\midrule
\multirow{7}{*}{OPT-6.7B} & bf16 & \textbackslash & \textbackslash &  100.00\%&30.38 & 65.53 & 65.72 & 50.53 & 76.22 & 64.88 & 24.94 & 12.28 \\
\cmidrule{2-13}
& & LLM-QAT & -1 & 12.5\%& 19.03 & 38.51 & 60.64 & 30.34 & 60.39 & 50.59 &22.92& 53.24 \\
& \multirow{6}{*}{2-bit} & BitDistiller & -1 & 12.5\% & 21.50 & 42.55 & 48.69 & 31.88 & 61.10 & 52.41 & 24.57 & 125.01 \\
& & \textbf{R2Q (ours)} & -1 & 12.5\%& \textbf{25.68} & \textbf{58.08} & \textbf{63.7} & \textbf{44.19} & \textbf{71.43} & \textbf{60.77} & \textbf{24.60} &\textbf{16.71} \\
\cmidrule{3-13}
& & LLM-QAT & 128 & 13.5\% &19.71 & 39.23 & 59.97 & 36.17 & 61.75 & 51.01 & \textbf{25.63} & 31.07 \\
& & BitDistiller & 128 &13.5\% & \textbf{26.37} & 53.70 & 62.35 & 41.22 & 68.66 & 56.99 & 25.40 & 22.49 \\
& & \textbf{R2Q (ours)} & 256 & 13.4\% & 24.32 & \textbf{58.96} & \textbf{64.40} & \textbf{44.31} & \textbf{72.58} & \textbf{60.54} & 25.09 & \textbf{16.81} \\
\bottomrule
\end{tabular}
}
}
\label{tbl:compare_experiment}
\end{table*}

\subsection{Experimental Configurations}\label{subsec:config}

In this section, we present the configurations of our experimental framework. We describe experimental settings, the generation strategy for training data and the evaluation benchmarks along with the associated metrics.

\textbf{Experimental Setup.} We adopt the knowledge distillation (KD) framework following LLM-QAT~\citep{liu2023llm-qat}. The BFloat16 model is used as the teacher, while the 2-bit quantized model serves as the student. The student is optimized to align with the teacher’s output distribution by minimizing the Kullback–Leibler (KL) divergence loss. Training is performed using the AdamW optimizer with a learning rate of $2\times10^{-5}$, zero weight decay, and a cosine learning rate schedule. The input sequence length is capped at 2048 tokens. We evaluate our approach on representative LLM architectures, including OPT~\citep{zhang2022opt}, Llama~\citep{touvron2023llama}, and Qwen~\citep{qwen2.5,yang2025qwen3}. All experiments are conducted on a single NVIDIA A800 GPU (80 GB), and each training session takes approximately 6 GPU hours.

\textbf{Training Data.} Following to the data-free paradigm of LLM-QAT, we utilize synthetic data generated by the teacher model for distillation. Each sequence is initiated by a uniformly sampled token, after which the teacher model generates subsequent tokens autoregressively. To establish a coherent context, the initial 3–5 tokens are selected deterministically (top-1 sampling). The remainder of the sequence is generated via stochastic sampling to enhance data diversity and mitigate potential mode collapse. We cap the maximum sequence length at 512 tokens to maintain a balance between contextual diversity and computational cost.

\textbf{Evaluation Benchmarks and Metrics.} We assess the performance of our method on a diverse set of tasks spanning commonsense reasoning, question answering, and language modeling. For question answering and commonsense reasoning tasks, accuracy serves as the primary evaluation metric. For language modeling, we report perplexity (PPL), where lower values indicate superior performance. Following established quantization methodology~\cite{du2024bitdistiller}, the full-precision bf16 model performance is employed as the baseline. The specific benchmarks include: \textbf{ARC}~\citep{clark2018arc}, which comprises science questions at both easy and challenge difficulty levels (ARC-e and ARC-c); \textbf{BoolQ}~\citep{clark2019boolq}, a binary (yes/no) question answering task based on paragraph entailment; and \textbf{HellaSwag}~\citep{zellers2019hellaswag}, which evaluates commonsense inference by tasking the model with selecting the most plausible continuation of a given context. Furthermore, \textbf{PIQA}~\citep{bisk2020piqa} focuses on physical commonsense reasoning between two provided options, while \textbf{Winogrande}~\citep{sakaguchi2021winogrande} assesses pronoun disambiguation capabilities that require nuanced contextual understanding. \textbf{MMLU}~\citep{hendrycks2020mmlu1}, evaluated under a 5-shot setting, provides a comprehensive measure of multi-domain knowledge and reasoning across a broad range of professional and academic subjects. Finally, for language modeling, we utilize \textbf{WikiText-2}~\citep{merity2016wikitext} to assess next-word prediction capabilities on a curated corpus of Wikipedia text.

\subsection{Evaluation on Language Understanding and Modeling Tasks}\label{sec:eval on benchmarks}
Table~\ref{tbl:compare_experiment} presents a comprehensive evaluation of our proposed R2Q method. We evaluated R2Q against the current state-of-the-art 2-bit quantization method, BitDistiller~\citep{du2024bitdistiller}. Consistent with BitDistiller’s approach, we incorporated LLM-QAT~\citep{liu2023llm-qat} as a comparative method, which utilizes knowledge distillation. We report performance across seven representative benchmarks using two foundation models: Llama-7B~\citep{touvron2023llama} and OPT-6.7B~\citep{zhang2022opt}. The compression ratio (CR) of the model's memory is presented.

For group-wise quantization, we consider coarse-grained and fine-grained settings. In coarse-grained settings, each channel forms a group (group size = -1). The fine-grained baseline uses a group size of 128. Since R2Q uses two 1-bit kernels with separate scaling factors, it doubles the scaling parameters. To ensure a fair comparison with a similar parameter budget, we set R2Q’s group size to 256. R2Q consistently outperforms baselines across all settings. In the coarse-grained setting, it yields substantial gains—for instance, on Llama-7B, R2Q boosts ARC-e from 26.81 (LLM-QAT) and 25.88 (BitDistiller) to 56.82, and reduces WikiText-2 perplexity from over 1700 and 15,938 to just 17.13. On OPT-6.7B, it improves HellaSwag accuracy from 30.34 to 44.19 and lowers perplexity from 53.24 to 16.71. In fine-grained settings, R2Q maintains its edge: with group size 256, it achieves top scores—58.96 on ARC-e and 72.58 on PIQA—and the lowest WikiText-2 perplexity (16.81), outperforming all baselines.

These results highlight R2Q’s robustness across both discriminative and generative tasks. By decomposing quantization into two 1-bit kernels with enhanced error correction, R2Q effectively preserves model capacity under aggressive compression. The effectiveness of this design is confirmed through ablation studies based solely on the initial 1-bit coarse quantization. Details are presented in Appendix~\ref{sec:ablation_residual}

\begin{table*}[h]
\centering
\caption{Performance comparison between the original BitDistiller (BitDistiller (RTN)) and BitDistiller integrated with R2Q (BitDistiller (R2Q)) under coarse-grained (group size of -1) quantization.}
{\setlength{\tabcolsep}{7pt}
\resizebox{\linewidth}{!}{ %
\begin{tabular}{llcccccccc}
\toprule
\textbf{Model} & \textbf{Method}  & \textbf{ARC-c}$\uparrow$ & \textbf{ARC-e}$\uparrow$ & \textbf{BoolQ} $\uparrow$ & \textbf{Hella.} $\uparrow$ & \textbf{PIQA}$\uparrow$ & \textbf{Wino.}$\uparrow$ & \textbf{MMLU}$\uparrow$& \textbf{Wiki2}$\downarrow$  \\
\midrule
\multirow{2}{*}{Llama-7B} & BitDistiller (RTN)& \textbf{23.38}&25.88&53.88& 25.59& 51.46&50.43& \textbf{24.61} &15938.61\\ 
& BitDistiller (\textbf{R2Q}) & 20.99 & \textbf{40.11}&\textbf{54.80}& \textbf{30.17} & \textbf{60.66}&\textbf{51.46}& 23.05 & \textbf{310.03} \\ 
\cmidrule{2-10} 
\multirow{2}{*}{OPT-6.7B} & BitDistiller (RTN) & 21.50 & 42.55 & 48.69& 31.88 & 61.10& 52.41&24.57 & 125.01\\ 
& BitDistiller (\textbf{R2Q}) & \textbf{25.09} & \textbf{55.01} & \textbf{63.36} & \textbf{41.30}  & \textbf{71.11} & \textbf{57.30}  &\textbf{24.68} &\textbf{112.55}    \\ 
\cmidrule{2-10} 
\multirow{2}{*}{Qwen2.5-7B} & BitDistiller (RTN)&20.39&23.76&42.20&25.02&52.45&49.64& 26.85 &42877.17\\ 
& BitDistiller (\textbf{R2Q}) & \textbf{36.09} & \textbf{61.61}&\textbf{70.03}& \textbf{39.59} & \textbf{68.99}&\textbf{60.69}& \textbf{43.88} &\textbf{193.05} \\ 
\cmidrule{2-10} 
\multirow{2}{*}{Qwen3-4B} & BitDistiller (RTN) &20.39&26.30&38.10&25.75&52.34&48.86& 24.48 &44696.35 \\ 
& BitDistiller (\textbf{R2Q}) & \textbf{23.63} & \textbf{43.77} & \textbf{66.17} & \textbf{31.72}  & \textbf{59.41} & \textbf{53.20}  & \textbf{29.52} & \textbf{494.6} \\ 
\bottomrule
\end{tabular}
}
}
\label{tbl:plug_and_play}
\end{table*}

\begin{table*}[h]
\centering
\caption{Benchmark-wise comparison of quantization stability for Llama-7B and OPT-6.7B with varying group sizes.The table reports the changes in model performance from coarse-grained to fine-grained quantization. The smaller the absolute value, the more stable the quantization strategy.}
{\setlength{\tabcolsep}{2pt} %
\resizebox{\linewidth}{!}{ %
\begin{tabular}{llccccccccccc}
\toprule
\textbf{Model}& \textbf{Method} & \textbf{Group-Size} & \textbf{CR} & $\Delta$\textbf{ARC-c} & $\Delta$\textbf{ARC-e} & $\Delta$\textbf{BoolQ} & $\Delta$\textbf{Hella.}  & $\Delta$\textbf{PIQA} & $\Delta$\textbf{Wino.}& $\Delta$\textbf{MMLU} & $\Delta$\textbf{Wiki2} \\
\midrule
\multirow{3}{*}{Llama-7B} & LLM-QAT & g128$\to$g-1 &15.1\%$\to$12.3\%& -6.74 & -15.19 & -20.49 & -9.61 & -10.23 & -3.94 &-0.96& 1661.46 \\
& BitDistiller & g128$\to$g-1 &15.1\%$\to$12.3\%& -5.29 & -32.03 & -9.30 & -16.17 & -17.64 & -9.79 & 0.35 & 15905.10 \\
& \textbf{R2Q (ours)} & g256$\to$g-1 &15.0\%$\to$12.3\%& \textbf{-0.77} & \textbf{-2.36} & \textbf{-5.53} & \textbf{-0.93} & \textbf{-0.54} & \textbf{-2.05} & \textbf{-0.20} & \textbf{0.17} \\
\midrule
\multirow{3}{*}{OPT-6.7B} & LLM-QAT & g128$\to$g-1  &13.5\%$\to$12.5\% & -0.68 & -0.72 & 0.67 & -5.83 & -1.36 & -0.42 &-2.71& 22.17 \\
& BitDistiller & g128$\to$g-1 &13.5\%$\to$12.5\%& -4.87 & -11.15 & -13.66 & -9.34 & -7.56 & -4.58 &-0.83& 102.52 \\
& \textbf{R2Q (ours)} & g256$\to$g-1 & 13.4\%$\to$12.5\%  & \textbf{1.36} & \textbf{-0.88} & \textbf{-0.70} & \textbf{-0.12} & \textbf{-1.15} & \textbf{0.23} & \textbf{-0.49} &\textbf{-0.10} \\
\bottomrule
\end{tabular}
}
}
\label{tbl:stability}
\end{table*}

\subsection{R2Q as a Plug-and-Play Module}
Beyond direct quantization performance, we further assess the adaptability of R2Q when integrated into existing quantization pipelines. To validate the versatility of R2Q, we integrate it into the BitDistiller framework under a coarse-grained quantization setting. Specifically, we compare BitDistiller using RTN (BitDistiller (RTN)) with its R2Q-enhanced counterpart (BitDistiller (R2Q)) across language understanding and modeling benchmarks on Llama-7B, OPT-6.7B, Qwen2.5-7B~\citep{qwen2.5} and Qwen3-4B~\citep{yang2025qwen3} (Table~\ref{tbl:plug_and_play}). Further details can be found in Appendix~\ref{sec:plug_and_play}.

Across multiple models, R2Q consistently demonstrates superior generalization under aggressive quantization. On Llama-7B, while BitDistiller (RTN) performs slightly better on ARC-c, R2Q integration yields substantial improvements elsewhere, including +14.23 on ARC-e, +4.58 on HellaSwag, and a sharp reduction in WikiText-2 perplexity. On OPT-6.7B, R2Q consistently outperforms RTN, boosting BoolQ, PIQA, and Winogrande, achieving +12.46 on ARC-e, and lowering WikiText-2 perplexity from 125.01 to 112.55. On Qwen2.5-7B, the improvements are even more striking, with +37.85 on ARC-e , +27.83 on BoolQ, +17.03 on MMLU and a drastic drop in WikiText-2 perplexity. On Qwen3-4B, R2Q integration restores performance exceptionally well, achieving +28.07 on BoolQ , +17.47 on ARC-e, + on 5.04 on MMLU and reducing WikiText-2 perplexity from 44,696.35 to 494.6. 

It is worth noting that MMLU scores for Llama-7B and OPT-6.7B are relatively low even in full precision, making their performance degradation at 2-bit quantization expected. In contrast, for Qwen2.5-7B and Qwen3-4B, BitDistiller with R2Q achieves significantly higher accuracy than with RTN—where RTN collapses to near-random guessing—underscoring the robustness of R2Q. Additional experiments and analyses are provided in the appendix.

These results highlight R2Q’s remarkable effectiveness in recovering and enhancing model performance after extremely coarse-grained and low-bit quantization, while maintaining or improving performance on challenging benchmarks such as MMLU.

\subsection{Quantization Stability}\label{sec:stability}
Table~\ref{tbl:stability} compares quantization stability under different group sizes, measured by the performance change ($\Delta$) across language understanding and modeling tasks. Smaller absolute values indicate better stability.

R2Q consistently outperforms BitDistiller and LLM-QAT across most tasks and group size transitions. For example, with Llama-7B from g256 to g-1, R2Q shows only -0.77\% on ARC-c, versus -5.29\% (BitDistiller) and -6.74\% (LLM-QAT). On BoolQ, R2Q has a -5.53\% drop, significantly lower than -9.3\% and -20.49\%. For OPT-6.7B, R2Q shows strong stability, even slight gains like +1.36\% on ARC-c, and near-zero drops on HellaSwag (-0.12\%) and WikiText-2 (-0.1\%).

These results validate R2Q effectively mitigates accuracy degradation during aggressive group-wise quantization, providing a robust choice for stable low-bit model compression.

\begin{figure}[!htb]
    \centering
    \includegraphics[width=1.0\linewidth,trim=0cm 5cm 0cm 5cm,clip]{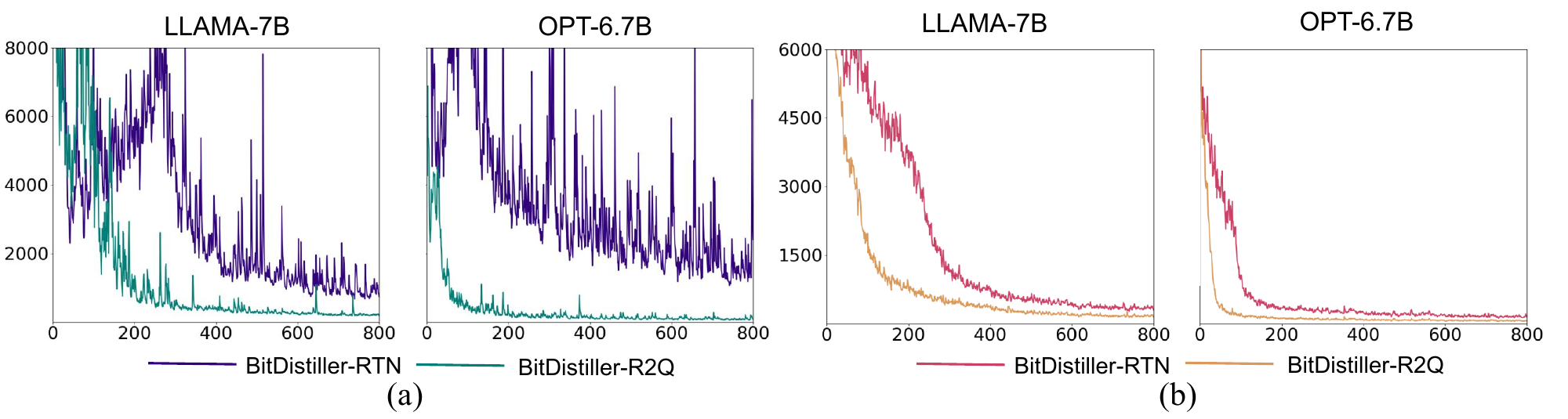}
    \caption{The (a) gradient norms and (b) training loss of BitDistiller (RTN) and BitDistiller (R2Q). The R2Q-integrated version (BitDistiller-R2Q) significantly reduces gradient fluctuations and converges faster and smoother.}
    \label{fig:grad_loss}
\end{figure}

\textbf{Gradient Stability and Convergence Efficiency.} Figure~\ref{fig:grad_loss} shows training gradient norms for Llama-7B and OPT-6.7B. Standard RTN causes unstable, oscillatory gradients, while R2Q integration yields smoother curves and faster convergence, indicating better optimization dynamics and less quantization noise. These results show R2Q can effectively serve as a plug-in module, enhancing robustness and performance across architectures and tasks in coarse-grained quantization.

\subsection{Quantization Error Analysis}\label{sec:quantization_error}
We analyze the source of R2Q’s effectiveness by measuring the quantization error before and after QAT. The Mean Squared Error (MSE) between original and quantized weights is computed as
\begin{equation}
\boldsymbol{E} = \frac{1}{L} \sum_{j=1}^{L} \frac{\|\mathbf{W}_j - \hat{\mathbf{W}}_j\|_2^2}{N_j},
\end{equation}
where $L$ is the number of linear layers in a decoder block and $N_j$ is the parameter count of layer $j$.

\begin{figure}
\centering
\includegraphics[width=1.0\linewidth,trim=4.2cm 5.2cm 3.5cm 5.2cm,clip]{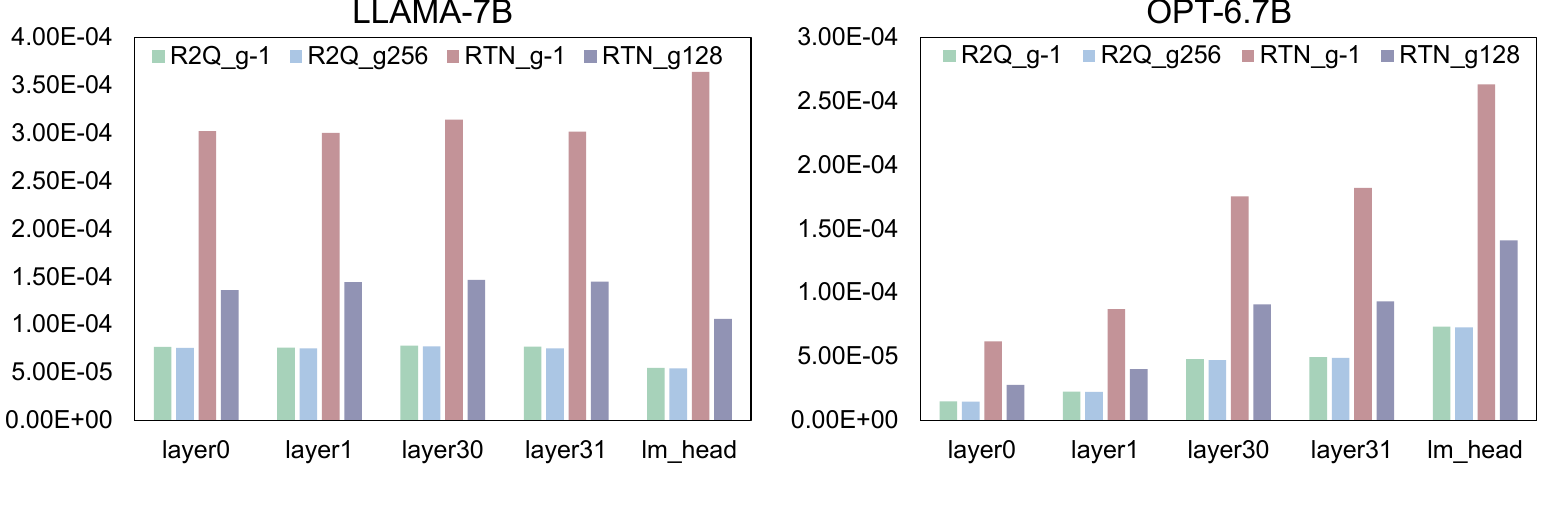}
\caption{Weight deviations before and after QAT under RTN and R2Q. MSE is used as the evaluation metric. For clarity, we report results for the first two decoder layers (layer0 and layer1), the last two decoder layers (layer30 and layer31), and the LM head of OPT-6.7B and Llama-7B. R2Q consistently maintains close alignment with the full-precision weights in both coarse-grained and fine-grained settings, whereas RTN shows weak alignment and suffers substantial degradation under coarse-grained quantization.}
\label{fig:error}
\end{figure}

Figure~\ref{fig:error} shows that RTN incurs substantially higher error under coarse granularity (RTN\_g-1) than fine granularity (RTN\_g128), often doubling the MSE across all inspected layers. This reflects RTN’s inefficient use of the quantization lattice, where a limited set of dominant values collapses onto few quantization points, reducing representational fidelity. R2Q, in contrast, is highly robust to granularity. The error gap between coarse (R2Q\_g-1) and fine (R2Q\_g256) settings remains small across all layers, consistent with the stability trends discussed in Section~\ref{sec:stability}. Even under extreme per-channel quantization, R2Q preserves the structural properties of the original weights.

Across both coarse-grained and fine-grained configurations, R2Q consistently attains lower quantization error than RTN, providing clear evidence of its superior quantization accuracy and explaining its performance advantages over RTN-based 2-bit methods.
\section{Discussion and Future Work}\label{sec:Discussion_and_Future_Work} 
We proposed Residual Refinement Quantization (R2Q), a novel framework that decomposes 2-bit quantization into two optimal, distribution-free 1-bit subproblems, providing greater flexibility than conventional approaches. Experimental results demonstrate that R2Q consistently improves task performance, enhances gradient stability, and accelerates convergence, particularly under coarse-grained quantization. As a plug-and-play module, R2Q can be seamlessly integrated into existing frameworks, delivering notable gains without any architectural modifications.

\textbf{Limitations and Future Work.} While R2Q effectively addresses weight quantization, it currently does not extend to activation quantization, which may limit its overall compression ratio and inference speedup. Extending R2Q to jointly quantize both weights and activations constitutes a promising yet challenging direction. Another limitation arises from the heuristic assumption that combining two 1-bit kernels yields an optimal, distribution-independent 2-bit representation. Although this assumption is supported empirically, a formal theoretical justification remains an open problem. Future work will focus on developing high-performance R2Q operators implemented with 1-bit arithmetic logic (see Appendix~\ref{sec:matmul}), which enables fully parallel execution of the two sub-kernels without sacrificing inference efficiency. Moreover, our investigation primarily focuses on the generalization effects of R2Q. Its potential for improving quantization-aware training (QAT) and fine-tuning remains largely unexplored. Extending R2Q to more diverse architectures, including multimodal and sparse transformers, may further reveal its adaptability and uncover architecture-specific quantization dynamics. Finally, exploring automated strategies for kernel selection and subproblem decomposition could further enhance R2Q’s efficiency and robustness.

In summary, R2Q establishes a flexible and extensible foundation for advancing ultra-low-bit quantization in large-scale models, paving the way for future research in both algorithmic innovation and system-level optimization.

{
    \small
    \bibliographystyle{ieeenat_fullname}
    \bibliography{cvpr26}
}

\clearpage
\setcounter{page}{1}
\maketitlesupplementary

\noindent\textbf{\Large Appendix}\\~\\
This appendix provides the supplementary materials for this work ``R2Q: Residual Refinement Quantization for Robust 2-Bit Large Language Models'', constructed according to the corresponding sections therein.

\begin{table*}[t]
\centering
\caption{Performance comparison between the original BitDistiller (BitDistiller-RTN) and the R2Q-enhanced BitDistiller (BitDistiller-R2Q) under both coarse-grained and fine-grained quantization. A dash (‘–’) indicates that training encountered gradient explosion and thus no valid metrics were produced.}
{\setlength{\tabcolsep}{2pt}
\resizebox{\linewidth}{!}{ %
\begin{tabular}{llccccccccc}
\toprule
\textbf{Model} & \textbf{Method} &\textbf{Group-Size} & \textbf{ARC-c}$\uparrow$ & \textbf{ARC-e}$\uparrow$ & \textbf{BoolQ} $\uparrow$ & \textbf{Hella.} $\uparrow$ & \textbf{PIQA}$\uparrow$ & \textbf{Wino.}$\uparrow$ & \textbf{MMLU}$\uparrow$ &\textbf{WikiText-2}$\downarrow$  \\
\midrule
\multirow{4}{*}{Llama-7B} &\multirow{2}{*}{BitDistiller (RTN)}& -1 & 23.38&25.88&53.88& 25.59& 51.46&50.43&24.61&15938.61\\ 
 && 128 &28.67&57.91&63.18&41.76&69.1&60.22&24.26&33.5056\\ 
  \cmidrule{3-11} 
& \multirow{2}{*}{BitDistiller (\textbf{R2Q})} & -1 & 20.99&40.11&54.80& 30.17&60.66&51.46&23.05&310.03\\
 && 256 &20.56&28.03&47.89&26.13&53.54&50.12&24.94&8104.6957\\ 
\cmidrule{2-11} 
\multirow{4}{*}{OPT-6.7B} &  \multirow{2}{*}{BitDistiller (RTN)}& -1 & 21.50 & 42.55 & 48.69& 31.88 & 61.10& 52.41&24.57& 125.01\\ 
 && 128 &26.37&53.7&62.35&41.22&68.66&56.99&25.40&22.49\\ 
 \cmidrule{3-11} 
& \multirow{2}{*}{BitDistiller (\textbf{R2Q})}& -1 &25.09&55.01&63.36&41.30&71.11&57.30&24.68&112.55\\ 
 && 256 &24.32&54.21&63.52&41.36&70.84&59.19&25.96&95.56\\ 
\cmidrule{2-11} 
\multirow{4}{*}{Qwen2.5-7B} &\multirow{2}{*}{BitDistiller (RTN)}& -1 &20.39&23.76&42.20&25.02&52.45&49.64&26.85&42877.17\\ 
 & & 128 &-&-&-&-&-&-&-&-\\ 
 \cmidrule{3-11} 
& \multirow{2}{*}{BitDistiller (\textbf{R2Q})} & -1 &36.09&61.61&70.03& 39.59 & 68.99&60.69&43.88&193.05 \\ 
& & 256 &36.09&63.80&68.78&40.64&70.67&62.43&48.85&122.18\\ 
\cmidrule{2-11} 
\multirow{4}{*}{Qwen3-4B} & \multirow{2}{*}{BitDistiller (RTN)} & -1 &20.39&26.30&38.10&25.75&52.34&48.86&24.48&44696.35 \\ 
 & & 128 &30.12&54.59&71.04&36.01&63.82&55.41&33.15&115.24\\ 
\cmidrule{3-11} 
&\multirow{2}{*}{BitDistiller (\textbf{R2Q})}& -1 &23.63&43.77&66.17&31.72&59.41&53.20&29.52&494.6\\ 
 & & 256 &32.68&63.68&72.51&38.35&64.52&55.8&44.07&61.93\\ 
\cmidrule{2-11} 
\multirow{4}{*}{Qwen3-8B}& \multirow{2}{*}{BitDistiller (RTN)}& -1 &-&-&-&-&-&-&-&-\\ 
 & & 128 &-&-&-&-&-&-&-&-\\ 
 \cmidrule{3-11} 
& \multirow{2}{*}{BitDistiller (\textbf{R2Q})}&-1&35.92&63.26&76.33&41.48&68.77&59.90&42.90&62.96\\ & & 256 &37.54&65.45&74.25&42.16&70.08&60.06&47.24&50.71\\ 

\bottomrule
\end{tabular}
}
}
\label{tbl:plug_and_play_apd}
\end{table*}
\section{LLM Usage Statement}
In the preparation of this manuscript, we utilized a large language model (LLM) to assist with language editing and refinement. Specifically, we used Google's Gemini Pro and OpenAI's GPT-4 for tasks such as correcting grammar, improving sentence structure for clarity, and ensuring consistency in terminology. Our process involved first drafting the content ourselves to articulate the core scientific ideas, methodology, and results. Subsequently, the LLM was prompted to polish the language of the human-authored text. All suggestions provided by the LLM were critically reviewed, edited, and approved by the authors to ensure the final text accurately reflects our original research and intent. The intellectual contribution, including all concepts, experimental design, analysis, and conclusions presented in this paper, is entirely the work of the human authors.

\section{R2Q as a Plug-and-Play Module}\label{sec:plug_and_play}
To demonstrate the efficacy and plug-and-play nature of R2Q, we integrated it into the BitDistiller post-training quantization (PTQ) framework. The primary results are summarized in Table~\ref{tbl:plug_and_play} (Section~\ref{sec:Experiments}), with detailed breakdowns provided in Table~\ref{tbl:plug_and_play_apd}.

\textbf{Llama-7B.} Under coarse-grained quantization (Group-Size = -1), BitDistiller (RTN) slightly outperforms R2Q on ARC-c (23.38 vs. 20.99). However, R2Q achieves substantial gains on ARC-e (40.11 vs. 25.88), HellaSwag (30.17 vs. 25.59), PIQA (60.66 vs. 51.46), and drastically lowers WikiText-2 perplexity (310.03 vs. 15,938.61), highlighting its ability to recover model performance under extreme quantization. Under finer-grained quantization (Group-Size 128/256), trends favor larger group sizes for RTN, but R2Q consistently maintains competitive results across tasks.

\textbf{OPT-6.7B.} Under coarse-grained quantization, R2Q demonstrates superior performance over RTN across nearly all evaluated metrics: ARC-c (25.09 vs. 21.50), ARC-e (55.01 vs. 42.55), BoolQ (63.36 vs. 48.69), HellaSwag (41.30 vs. 31.88), PIQA (71.11 vs. 61.10), and Winogrande (57.30 vs. 52.41), while also reducing WikiText-2 perplexity from 125.01 to 112.55. In the fine-grained setting (Group-Size 256), R2Q continues to either match or slightly surpass RTN, highlighting its robust generalization capabilities.

\textbf{Qwen2.5-7B.} The performance enhancements afforded by R2Q are particularly pronounced for Qwen2.5-7B under coarse-grained quantization. For instance, the score on ARC-e improves from 23.76 to 61.61, BoolQ from 42.20 to 70.03, and WikiText-2 perplexity is drastically reduced from 42,877.17 to 193.05. Notably, the baseline RTN method experienced gradient explosion during training in fine-grained settings, resulting in no reported metrics (denoted by '-'). This observation underscores the enhanced training stability provided by R2Q. With fine-grained quantization (Group-Size 256), R2Q maintains its strong performance across all benchmarks.

\textbf{Qwen3-4B.} Similarly, integrating R2Q significantly enhances the performance of Qwen3-4B under coarse quantization settings. Performance on BoolQ increases from 38.10 to 66.17, ARC-e from 26.30 to 43.77, and WikiText-2 perplexity decreases from 44,696.35 to 494.6. Under a fine-grained configuration (Group-Size 256), R2Q continues to match or exceed the performance of RTN, achieving scores of 63.68 on ARC-e and 72.51 on BoolQ.

\textbf{Qwen3-8B.} The results on Qwen3-8B conclusively demonstrate the critical advantage of our approach. The baseline BitDistiller (RTN) failed completely in both coarse and fine-grained settings due to severe gradient explosion, rendering the model untrainable. In sharp contrast, R2Q exhibited exceptional stability, successfully converging to achieve state-of-the-art (SOTA) performance for 2-bit quantization. Even under coarse settings, R2Q achieved impressive capabilities, with a BoolQ score of 76.33 and a WikiText-2 perplexity of 62.96. Fine-grained quantization (Group-Size 256) further elevated performance, reaching 47.24 on MMLU and 70.08 on PIQA, validating R2Q as the most robust solution for quantizing modern, high-performance architectures.

It is crucial to highlight that R2Q offers superior stability compared to standard RTN-based QAT. Specifically, for Qwen2.5-7B and Qwen3-8B, the \textbf{RTN method encountered gradient explosion}, leading to training failure. R2Q effectively mitigates these issues, remaining stable across all tested models while delivering significant performance gains. These experiments validate that R2Q is a highly effective, stable, and versatile plug-and-play solution that establishes a new state-of-the-art for the 2-bit quantization of large language models.

\begin{table*}[t]
\centering
\caption{Ablation study on the contribution of the residual refinement stage of R2Q. The initial 1-bit coarse approximation against the complete 2-bit R2Q method is compared over language understanding and modeling tasks. For ARC-c/e, BoolQ, Hellaswag, PIQA, and Winogrande, we report accuracy. For WikiText-2, we
report PPL. To align the scale parameters of the 1-bit coarse-estimated control group with the 2-bit full R2Q method, we set the group size of 128 for the former and 256 for the latter in fine-grained quantification.}
\setlength{\tabcolsep}{4pt}
\resizebox{\linewidth}{!}{ %
\begin{tabular}{ccccccccccc}
\toprule
\textbf{Model} & \textbf{Bit-Width} & \textbf{Group-Size} & \textbf{ARC-c}$\uparrow$ & \textbf{ARC-e}$\uparrow$ & \textbf{BoolQ}$\uparrow$ & \textbf{Hella.}$\uparrow$ & \textbf{PIQA}$\uparrow$ & \textbf{Wino.}$\uparrow$ & \textbf{MMLU}$\uparrow$ & \textbf{Wiki2}$\downarrow$ \\
\midrule
\multirow{5}{*}{Llama-7B} & bf16 & \textbackslash & 41.98 & 75.38 & 75.17 & 56.95 & 78.78 & 69.46 & 31.33 &9.39 \\
\cmidrule{2-11}
& 1-bit & -1 & 19.45 & 36.07 & 61.01 & 29.00 & 57.02 & 52.09 & 22.91 & 66.28 \\
& 2-bit & -1 & \textbf{27.47} & \textbf{56.82} & \textbf{59.36} & \textbf{44.44} & \textbf{70.08} & \textbf{57.54} & \textbf{24.08} &\textbf{17.13} \\
\cmidrule{2-11}
& 1-bit & 128 & 18.43 & 37.33 & 60.31 & 29.86 & 59.30 & 51.70 & 22.92 & 49.81 \\
& 2-bit & 256 & \textbf{28.24} & \textbf{59.18} & \textbf{64.89} & \textbf{45.37} & \textbf{70.62} & \textbf{59.59} & \textbf{24.28} & \textbf{16.96} \\
\midrule
\midrule
\multirow{5}{*}{OPT-6.7B} & bf16 & \textbackslash & 30.38 & 65.53 & 65.72 & 50.53 & 76.22 & 64.88 & 24.94 & 12.28 \\
\cmidrule{2-11}
& 1-bit & -1 & 20.73 & 40.45 & 53.06 & 30.11 & 60.28 & 50.51 &22.99 & 417.53 \\
& 2-bit & -1 & \textbf{25.68} & \textbf{58.08} & \textbf{63.70} & \textbf{44.19} & \textbf{71.43} & \textbf{60.77} & \textbf{24.60} & \textbf{16.71} \\
\cmidrule{2-11}
& 1-bit & 128 & 21.16 & 41.83 & 54.22 & 30.25 & 61.70 & 50.75 & 22.95 & 480.08 \\
& 2-bit & 256 & \textbf{24.32} & \textbf{58.96} & \textbf{64.40} & \textbf{44.31} & \textbf{72.58} & \textbf{60.54} & \textbf{25.09} & \textbf{16.81} \\
\bottomrule
\end{tabular}
}
\label{tbl:ablation_of_residual_refinement}
\end{table*}

\section{Ablation Study of the Residual Refinement}
\label{sec:ablation_residual}

\begin{table*}[!htb]
\centering
\caption{Performance comparison with state-of-the-art 2-bit models trained from scratch (BitNet b1.58 and $i$Fairy). Notably, R2Q achieves superior performance by fine-tuning existing models, whereas the baselines necessitate training from scratch. Bold indicates the highest score, and underline indicates the second-highest.}
{\setlength{\tabcolsep}{2pt}
\resizebox{\linewidth}{!}{ %
\begin{tabular}{lccccccccccc}
\toprule
\textbf{Model} &\textbf{Model-Size}&\textbf{Bit-Width}& \textbf{ARC-c}$\uparrow$ & \textbf{ARC-e}$\uparrow$ & \textbf{BoolQ} $\uparrow$ & \textbf{Hella.} $\uparrow$ & \textbf{PIQA}$\uparrow$ & \textbf{Wino.}$\uparrow$ & \textbf{OQ}$\uparrow$ &\textbf{AVG}$\uparrow$  \\
\midrule
BitNet b1.58 &1.3B&1.58-bit&24.20&54.90&56.70&37.70&68.80&55.80&19.60&45.39\\
$i$Fairy &1.3B& 2-bit &\underline{24.66}&\underline{56.65}&59.60&\textbf{38.69}&\textbf{69.80}&54.06&\textbf{22.20}&\underline{46.52}\\ 
\textbf{OPT-IML-Max (R2Q)}&1.3B&2-bit&24.15&52.44&\textbf{69.82}&34.34&66.32&\underline{55.88}&\underline{20.00}&46.14\\ 
\textbf{phi-1.5 (R2Q)}&1.3B&2-bit&\textbf{30.89}&\textbf{58.67}&\underline{63.58}&\underline{36.60}&\underline{68.82}&\textbf{59.43}&19.80&\textbf{48.26}\\ 
\bottomrule
\end{tabular}
}
}
\label{tbl:compar_ifairy}
\end{table*}
As described in Section~\ref{sec:Methodology}, our R2Q method adopts a two-step quantization strategy designed to balance efficiency with accuracy. The first step generates a 1-bit kernel, $\alpha_1\mathbf{Q}_1$, which serves as a \textbf{coarse approximation} of the full-precision weights $\mathbf{W}$. In the second step, R2Q quantifies the residual error $\mathbf{R} = \mathbf{W} - \alpha_1\mathbf{Q}_1$ and encodes it using another 1-bit kernel—constituting the \textbf{residual refinement} stage. This decomposition is central to R2Q's design, enabling it to go beyond naïve low-bit quantization methods by explicitly modeling and correcting quantization errors.

Unlike standard 2-bit quantization methods, such as Round-To-Nearest (RTN), which directly map full-precision weights to a small discrete set, R2Q introduces a hierarchical decomposition: a coarse 1-bit quantization followed by an explicit residual correction. This framework enhances traditional 1-bit quantization methods~\citep{wang:bitnet,xu2024onebit} by addressing their key limitation—uncompensated approximation errors—through structured refinement.

While Section~\ref{sec:Experiments} confirms the effectiveness of R2Q overall, we further conducted an ablation study to assess the independent contribution of the setting of residual refinement. Specifically, we compared the performance of (1) the initial 1-bit coarse approximation and (2) the full 2-bit R2Q method. The dequantization for the 1-bit coarse quantization is,

\begin{equation}
\hat{\boldsymbol{w}}^{(i)} = \alpha_1^{(i)}\boldsymbol{q}_1^{(i)}
\end{equation}

All evaluations were performed under the same experimental configurations described in Section~\ref{subsec:config}. The results are presented in Table~\ref{tbl:ablation_of_residual_refinement}. To ensure a fair comparison in terms of parameter count, we align the number of scaling factors by setting the group size of the 1-bit ablation baseline to 128 and that of the complete 2-bit R2Q model to 256 in the fine-grained quantization setting.

Across Llama-7B and OPT-6.7B, moving from 1-bit to 2-bit quantization yields substantial gains. On ARC-e, R2Q improves accuracy from 36.07$\to$56.82 (Llama-7B) and 40.45$\to$58.08 (OPT-6.7B), with relative gains of 57.5\% and 43.6\%. The largest benefit appears in language modeling: 1-bit models show extremely high perplexity—66.28 and 417.53—signaling severe degradation, while 2-bit R2Q reduces perplexity to 17.13 and 16.71, close to full-precision (9.39 and 12.28). This >90\% reduction underscores the critical role of the refinement bit in preserving generative capacity. Improvements persist under grouped settings: with group size 128 vs. 256, ARC-e rises from 37.33$\to$59.18 (Llama-7B) and 41.83$\to$58.96 (OPT-6.7B), confirming robustness under deployment constraints. Moreover, the refinement boosts not just individual benchmarks but sustains balanced performance across commonsense reasoning, reading comprehension, and causal reasoning, suggesting the second bit captures generalized error patterns.

These results strongly validate our central hypothesis: while a 1-bit kernel offers a compact but rough estimate of the weight distribution, it fails to retain the finer-grained information necessary for complex tasks. The second bit in R2Q serves as an adaptive corrective mechanism, effectively modeling residual errors and yielding a much closer approximation to the full-precision model. The ablation confirms that residual refinement is not just a marginal enhancement—it is a critical design component that enables R2Q to achieve high accuracy under extreme quantization.

\begin{table}[!htb]
\centering
\caption{Comparison of training resources and data requirements. R2Q demonstrates significantly higher efficiency, requiring orders of magnitude less data and computing power compared to the training-from-scratch baseline.}
{\setlength{\tabcolsep}{2pt}
\resizebox{\linewidth}{!}{ %
\begin{tabular}{lcccc}
\toprule
\textbf{Method} & \textbf{GPUs} & \textbf{Data-Size} & \textbf{Time-Cost} \\
\midrule
$i$Fairy & 32 NVIDIA H800&100B& - \\ 
\textbf{OPT-IML-Max (R2Q)} &8 NVIDIA 4090&7.7M& 1.5h\\ 
\textbf{phi-1.5 (R2Q)}&8 NVIDIA 4090&7.8M& 1.5h\\ 
\bottomrule
\end{tabular}
}
}
\label{tbl:compar_ifairy_resource}
\end{table}

\begin{table*}[t]
\centering
\caption{Comparison of 2-bit Qwen-8B against 16-bit models with equivalent memory footprints.}
{\setlength{\tabcolsep}{4pt}
\resizebox{\linewidth}{!}{ %
\begin{tabular}{lcccccccccc}
\toprule
\textbf{Model} &\textbf{Bit-Width} & \textbf{ARC-c}$\uparrow$ & \textbf{ARC-e}$\uparrow$ & \textbf{BoolQ} $\uparrow$ & \textbf{Hella.} $\uparrow$ & \textbf{PIQA}$\uparrow$ & \textbf{Wino.}$\uparrow$ & \textbf{MMLU}$\uparrow$ &\textbf{WikiText-2}$\downarrow$  \\
\midrule
Qwen2.5-0.5B&bf16&29.10&64.81&62.08&40.58&70.40&56.35&47.51&17.60\\
Qwen2.5-1.5B&bf16&41.55&75.34&72.78&50.20&75.95&63.06&59.70&12.09\\
Qwen3-0.6B &bf16&31.83&60.81&64.31&37.54&67.25&56.82&40.07&26.13\\
Qwen3-1.7B &bf16&39.85&72.55&77.58&46.18&72.42&61.09&55.61&21.04\\
\textbf{Qwen3-8B (R2Q)} & 2-bit &37.12&66.96&77.83&43.46&69.70&59.98&46.39&32.46\\ 
\bottomrule
\end{tabular}
}
}
\label{tbl:same_scale_compare}
\end{table*}

\section{Comparison with State-of-the-Art 2-Bit Training-from-Scratch Models}\label{sec:compare_with_ifairy}
To further validate the effectiveness of R2Q, we compared it against state-of-the-art 2-bit methods: BitNet b1.58~\citep{wang:bitnet} and $i$Fairy~\citep{wang2025ifairy}. The results are presented in Table~\ref{tbl:compar_ifairy}.

It is crucial to note the fundamental difference in methodology: both BitNet b1.58 and $i$Fairy are training-from-scratch approaches, requiring massive computational resources and data to train 2-bit weights from initialization. Table~\ref{tbl:compar_ifairy_resource} provides a quantitative comparison of these resource requirements. While $i$Fairy necessitates a cluster of 32 NVIDIA H800 GPUs and a massive training corpus of 100B tokens, R2Q demonstrates extreme efficiency. Utilizing only 8 consumer-grade NVIDIA RTX 4090 GPUs and less than 8M tokens, R2Q completes the training in just 1.5 hours. In contrast, R2Q operates as a fine-tuning framework, transforming existing full-precision Large Language Models (LLMs) into 2-bit counterparts. Distinct from the data generation pipeline described in Section~\ref{subsec:config}, the fine-tuning data for this experiment was sourced from WikiText~\citep{merity2016wikitext}, Alpaca~\citep{alpaca}, and RedPajama~\citep{weber2024redpajama}. We randomly selected $5\text{k}$ samples from each dataset, resulting in a total of $15\text{k}$ samples used for fine-tuning.

Despite the theoretical advantage usually attributed to training from scratch, R2Q demonstrates superior or highly competitive performance. As shown in Table~\ref{tbl:compar_ifairy}, when applying R2Q to the phi-1.5 (1.3B) model, we achieve a new state-of-the-art average score of 48.26. This surpasses the strongest baseline, $i$Fairy (46.52), by 1.74 points, and outperforms BitNet b1.58 (45.39) by nearly 3 points. Specifically, the R2Q-quantized phi-1.5 shows distinct strengths in reasoning tasks, achieving significant leads in ARC-c (30.89 vs. 24.66) and Winogrande (59.43 vs. 54.06).

Regarding the OPT-IML-Max (1.3B), R2Q achieves an average score of 46.14. While slightly lower than the $i$Fairy average, it successfully outperforms BitNet b1.58. More importantly, the OPT-IML-Max model demonstrates exceptional performance on the BoolQ dataset with a score of 69.82, exceeding the $i$Fairy baseline (59.60) by over 10 points. This suggests that R2Q is particularly effective at preserving specific reasoning capabilities present in the parent model, even at extreme quantization levels.

These results challenge the prevailing assumption that effective 2-bit LLMs require training from scratch. We demonstrate that R2Q can effectively leverage the pre-trained knowledge of full-precision models to achieve performance comparable to, and in the case of phi-1.5, superior to state-of-the-art training-from-scratch methods, all while incurring significantly lower training costs as highlighted in Table~\ref{tbl:compar_ifairy_resource}.

\begin{table*}[h!]
\centering
\caption{A comparison of the computational complexity for a matrix multiplication operation ($M \times N$ by $N \times K$) across different quantization schemes. R2Q significantly reduces the number of floating-point multiplications required.}
\label{tbl:complexity_comparison}
\resizebox{0.9\linewidth}{!}{%
\begin{tabular}{cccc}
\toprule
&FP16&INT2&R2Q\\
\midrule
Mul&$MNK$ FP16$\times$FP16& $MNK$ INT2$\times$FP16 $+$ $MN$ FP16 & $2MN$ FP16$\times$FP16\\
\midrule
Add&$MN(K-1)$ FP16& $MN(K-1)$ FP16 & $2MN(K-1)$ FP16 $+$ $MN$ FP16 \\
\bottomrule
\end{tabular}
}
\end{table*}

\section{Performance Comparison with Memory-Equivalent Baselines}\label{sec:same_scale_compare}
To demonstrate the effectiveness of R2Q, we applied the method to quantize the Qwen3-8B model into 2 bits. Theoretically, a 2-bit model with $8\text{B}$ parameters has a memory footprint equivalent to a 16-bit model with $1\text{B}$ parameters. Therefore, to strictly evaluate the performance of the quantized model against models of comparable memory usage, we sought to compare it against a $1\text{B}$ parameter baseline. Since there is no available Qwen model with $1\text{B}$ parameters, we selected the Qwen2.5 series models ($0.5\text{B}$ and $1.5\text{B}$) and Qwen3 series models ($0.6\text{B}$ and $1.7\text{B}$) for comparison.

The quantization process utilized a diverse calibration set to ensure robustness. The data used for training includes C4~\citep{raffel2020c4}, Wikitext~\citep{merity2016wikitext}, FineWeb~\citep{penedo2024fineweb}, SlimPajama~\citep{weber2024redpajama}, OpenWebText~\citep{Gokaslan2019openweb}, Alpaca~\citep{alpaca}, Flan~\citep{longpre2023flan}, and OpenOrca~\citep{mukherjee2023openorca} datasets. From each of these datasets, $3\text{k}$ samples were randomly selected, resulting in a total of $24\text{k}$ samples (13.2M tokens) used for the calibration of  R2Q.

Table \ref{tbl:same_scale_compare} presents the performance comparison between the 2-bit Qwen3-8B (R2Q) and the bf16 baselines across various zero-shot benchmarks. The results indicate that our 2-bit Qwen8B model consistently outperforms the smaller baselines, specifically Qwen2.5-0.5B and Qwen3-0.6B. For instance, on the BoolQ benchmark, the 2-bit Qwen8B achieves a score of $77.83$, significantly surpassing the Qwen2.5-0.5B ($62.08$) and the Qwen3-0.6B ($64.31$). Similarly, in the ARC-c task, the quantized model scores $37.12$, outperforming the $29.10$ and $31.83$ scores of the 0.5B and 0.6B models, respectively.

Furthermore, despite the extreme compression to 2 bits, the model's performance is comparable to the larger Qwen3-1.7B model within the same series. While the Qwen3-1.7B achieves slightly higher scores on tasks like MMLU ($55.61$ vs $46.39$), the 2-bit Qwen8B actually outperforms the 1.7B model on BoolQ ($77.83$ vs $77.58$) and maintains competitive performance on HellaSwag ($43.46$ vs $46.18$) and WinoGrande ($59.98$ vs $61.09$). This demonstrates that R2Q allows an 8B parameter model to be compressed to the size of a $\approx 1\text{B}$ model while retaining capabilities often exceeding those of standard models with similar memory requirements.

\section{Matrix Multiplication for R2Q}\label{sec:matmul}

As detailed in Section~\ref{sec:Methodology}, R2Q represents a quantized weight matrix, $\hat{\mathbf{W}}\in \mathbb{R}^{M \times K}$, using two 1-bit kernels, $\mathbf{Q}_1$ and $\mathbf{Q}_2$, along with their corresponding scaling factors, $\boldsymbol{\alpha}_1$ and $\boldsymbol{\alpha}_2$. The reconstruction of the weight matrix is given by
\begin{equation}
\hat{\mathbf{W}} = \boldsymbol{\alpha}_1\mathbf{Q}_1 + \boldsymbol{\alpha}_2\mathbf{Q}_2
\end{equation}

This decomposition allows for a significant optimization of the matrix multiplication (Matmul) operation. For a given input matrix $\mathbf{X}\in \mathbb{R}^{K \times N}$, the product $\hat{\mathbf{W}}\mathbf{X}$ can be calculated by leveraging the distributive property
\begin{equation}
\hat{\mathbf{W}}\mathbf{X} = (\boldsymbol{\alpha}_1\mathbf{Q}_1 + \boldsymbol{\alpha}_2\mathbf{Q}_2)\mathbf{X} = \boldsymbol{\alpha}_1(\mathbf{Q}_1\mathbf{X}) + \boldsymbol{\alpha}_2(\mathbf{Q}_2\mathbf{X}) \label{eq:matmul_optimized}
\end{equation}

The reformulation in Equation~\ref{eq:matmul_optimized} breaks the computation into two highly efficient steps.

\begin{enumerate}
    \item \textbf{Binary Matrix Multiplication}: First, the products of the 1-bit matrices and the input matrix, $\mathbf{Q}_1\mathbf{X}$ and $\mathbf{Q}_2\mathbf{X}$, are computed. Since $\mathbf{Q}_1$ and $\mathbf{Q}_2$ contain only values of $+1$ and $-1$, this operation \textbf{eliminates the need for multiplications}. Instead, it is executed using only additions.

    \item \textbf{Scaling and Combination}: The resulting matrices from the first step are then scaled element-wise (Hadamard product) by their respective scaling factors, $\boldsymbol{\alpha}_1$ and $\boldsymbol{\alpha}_2$. Finally, the two scaled matrices are added together to produce the final result.
\end{enumerate}

Additionally, since the calculations of $\boldsymbol{\alpha}_1\mathbf{Q}_1\mathbf{X}$ and $\boldsymbol{\alpha}_2\mathbf{Q}_2\mathbf{X}$ are completely unrelated, both can be performed \textbf{in parallel}. This approach dramatically reduces the number of expensive multiplication operations. Table~\ref{tbl:complexity_comparison} provides a comparison of the computational complexity for matrix multiplication using full-precision (FP16), standard 2-bit integer (INT2), and our proposed R2Q method. As shown, R2Q substantially decreases the reliance on high-precision multiplications compared to the other methods.

\end{document}